%% file: rsi_pdebench.tex
\documentclass{article}
\ifdefined\pdfinclusioncopyfonts\pdfinclusioncopyfonts=1\fi

\usepackage{arxiv}

\usepackage[T1]{fontenc}
\usepackage[utf8]{inputenc}
\usepackage{amsmath}
\usepackage{charter}               
\usepackage[charter]{newtxmath}    
\usepackage{microtype}
\usepackage{graphicx}
\usepackage{subcaption}
\usepackage{booktabs}
\usepackage{multirow}
\usepackage{float}
\usepackage{url}
\usepackage{xcolor}
\definecolor{mydarkblue}{rgb}{0,0.08,0.45}
\definecolor{linkblue}{HTML}{1F66C9}
\usepackage[colorlinks=true,linkcolor=mydarkblue,citecolor=mydarkblue,urlcolor=linkblue,filecolor=mydarkblue]{hyperref}
\usepackage{natbib}
\setcitestyle{authoryear,round,citesep={;},aysep={,},yysep={;}}

\usepackage{mathtools}
\usepackage{tikz}
\usetikzlibrary{arrows.meta,positioning,fit,backgrounds,calc}
\usepackage{algorithm}
\usepackage{algorithmic}
\usepackage{fontawesome5}
\usepackage[capitalize,noabbrev]{cleveref}
\usepackage{placeins}
\usepackage[bottom]{footmisc}  
\usepackage{caption}
\DeclareCaptionLabelSeparator{bar}{\ \textbar\ }
\definecolor{rsired}{HTML}{D62828}
\definecolor{natblack}{HTML}{111111}
\definecolor{boxgray}{HTML}{F2F2F2}
\definecolor{evalgray}{HTML}{555555}

\newcommand{\bench}{RSI-PDEbench}
\newcommand{\rsipde}{RSI-PDE}
\newcommand{\tol}{\tau}

\makeatletter
\renewcommand\@seccntformat[1]{\csname the#1\endcsname.\quad}
\makeatother
\renewcommand{\shorttitle}{RSI-PDEbench}
\fancypagestyle{firstpage}{\fancyhf{}\cfoot{\thepage}}
\renewcommand{\footnoterule}{\kern-3pt\hrule width\textwidth height 0.4pt\kern 2.6pt}

\newcommand{\rsilogo}{%
  \begin{tikzpicture}
    \begin{scope}
      \clip (0,0) circle (0.31);
      \node at (0,0) {\includegraphics[width=0.66cm]{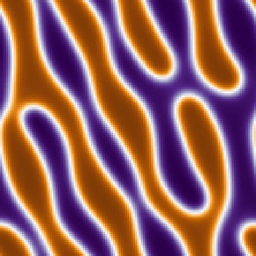}};
    \end{scope}
    \draw[rsired, line width=1.3pt, -{Stealth[length=3.4pt,width=4.4pt]}]
      (100:0.43) arc[start angle=100, end angle=-220, radius=0.43];
  \end{tikzpicture}}

\hypersetup{
  pdftitle={RSI-PDEbench: A Benchmark for Self-Improving Agents in PDE Solving},
  pdfauthor={Binghang Lu, Lang Cao, Changhong Mou, Yige Sun, Cindy Xiangrui Kong, Zheyuan Deng, Tian Yuan, Bing Hu, Runyu Zhang,
    Na Lu, Guang Lin},
  pdfkeywords={benchmark, self-improvement, coding agents, partial differential equations, scientific computing},
}

\title{RSI-PDEbench: A Benchmark for Self-Improving Agents in PDE Solving}
\author{Binghang~Lu\textsuperscript{1,*}, Lang~Cao\textsuperscript{2,\dag}, Changhong~Mou\textsuperscript{3,\dag},
    Yige~Sun\textsuperscript{4}, Cindy Xiangrui Kong\textsuperscript{1}, Zheyuan~Deng\textsuperscript{5},
    Tian~Yuan\textsuperscript{4}, Bing~Hu\textsuperscript{4}, Runyu~Zhang\textsuperscript{6},
    Na~Lu\textsuperscript{7} and Guang~Lin\textsuperscript{1,*}
    \\
    {\small \mbox{\textsuperscript{1}Purdue University}, \mbox{\textsuperscript{2}University of Illinois Urbana-Champaign},
    \mbox{\textsuperscript{3}Utah State University}, \mbox{\textsuperscript{4}Independent researcher},
    \mbox{\textsuperscript{5}Brown University}, \mbox{\textsuperscript{6}Massachusetts Institute of Technology},
    \mbox{\textsuperscript{7}Arizona State University}}
    \\
    {\small \textsuperscript{*}\textit{Corresponding authors:}
  \href{mailto:lu895@purdue.edu}{\textit{lu895@purdue.edu}}.\qquad \textsuperscript{\dag}\textit{Equal contribution.}}
}
\date{}

\renewcommand{\maketitle}{%
\begin{flushleft}
  {\fontsize{21}{26}\selectfont\bfseries
    \raisebox{-0.36em}{\rsilogo}\hspace{0.35em}\textcolor{rsired}{RSI-PDEbench}: A Benchmark for Self-Improving
    Agents in PDE Solving\par}
  \vspace{16pt}
  {\fontsize{10.5}{15}\selectfont\bfseries
    Binghang~Lu\textsuperscript{1,*}, Lang~Cao\textsuperscript{2,\dag}, Changhong~Mou\textsuperscript{3,\dag},
    Yige~Sun\textsuperscript{4}, Cindy Xiangrui Kong\textsuperscript{1}, Zheyuan~Deng\textsuperscript{5},
    Tian~Yuan\textsuperscript{4}, Bing~Hu\textsuperscript{4}, Runyu~Zhang\textsuperscript{6},
    Na~Lu\textsuperscript{7} and Guang~Lin\textsuperscript{1,*}\par}
  \vspace{4pt}
  {\small \mbox{\textsuperscript{1}Purdue University}, \mbox{\textsuperscript{2}University of Illinois Urbana-Champaign},
    \mbox{\textsuperscript{3}Utah State University}, \mbox{\textsuperscript{4}Independent researcher},
    \mbox{\textsuperscript{5}Brown University}, \mbox{\textsuperscript{6}Massachusetts Institute of Technology},
    \mbox{\textsuperscript{7}Arizona State University}\par}
\end{flushleft}
{\let\thefootnote\relax\footnotetext{\textsuperscript{*}\textit{Corresponding authors:}
  \href{mailto:lu895@purdue.edu}{\textit{lu895@purdue.edu}}.\qquad \textsuperscript{\dag}\textit{Equal contribution.}}}
\vspace{18pt}%
}

\begin{document}

\thispagestyle{firstpage}

\maketitle

{\bfseries\boldmath
Identifying and solving partial differential equations (PDEs) from sampled solution data often requires substantial expert effort in equation discovery, coefficient estimation, and numerical solver configuration, especially under noise and distribution shifts. Can large language model (LLM) agents automate this workflow accurately and efficiently? To investigate this question, we introduce \bench{}, a benchmark of 1,001 cases spanning 42 equation templates, including both classical and fractional PDEs. It provides three evaluation settings: Full evaluates performance on noise-free solution data; Noisy tests robustness to Gaussian noise with an RMS equal to 50\% of the clean-signal RMS; and Hard-OOD tests challenging out-of-distribution generalization to unseen equation templates and shifted coefficient ranges. We further introduce \rsipde{}, a recursively self-improving agent harness that shortlists candidate equations by residual loss, uses relative $L^2$ feedback to refine coefficients and numerical solvers, and applies a validation gate to policy and memory updates. We compare \rsipde{} with single-run baselines and native agent harnesses (Claude Code or Codex) across four frontier models. Across the three settings, \rsipde{} improves the pass rate averaged across the four models by 106.5--181.9\% relative to single-run baselines and by 3.0--19.6\% relative to native harnesses, while reducing total recorded model calls by 46.9\% across the 12 model--setting pairs. These results highlight the potential of validated recursive self-improvement in LLM agents to advance scientific computing.

\par}

\vspace{10pt}
\noindent{\small
  \href{https://github.com/jeremylu916/RSI-PDE}{\faGithub\ github.com/jeremylu916/RSI-PDE}\hfill
  \href{https://jeremylu916.github.io/RSI-PDE/}{\faGlobe\ jeremylu916.github.io/RSI-PDE}\hfill
  \href{https://huggingface.co/datasets/JeremyLu1017/PDE-RSI}{\raisebox{-0.22em}{\includegraphics[height=1.15em]{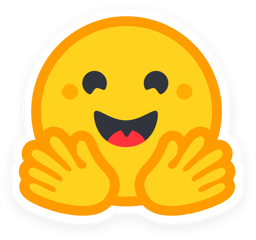}}\
    JeremyLu1017/PDE-RSI}\par}

\begin{figure}[H]
  \centering
  \includegraphics[width=\textwidth]{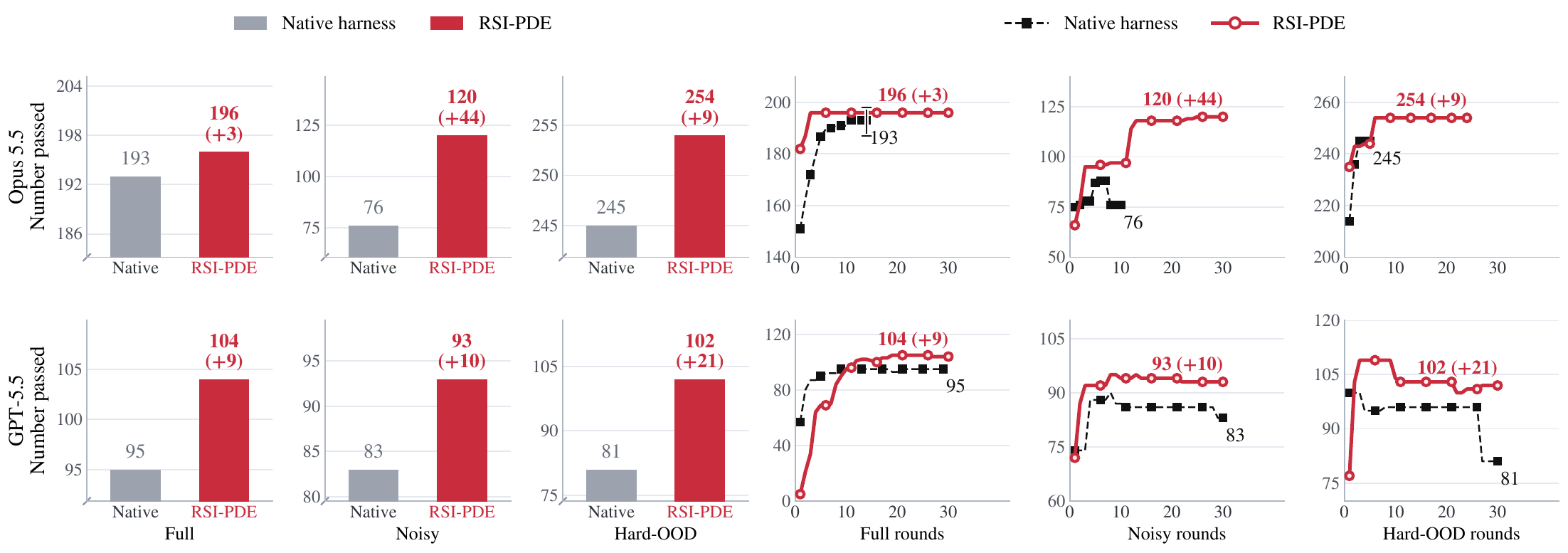}
  \caption{\textbf{Final accuracy and learning curves.} Left: passed-case bars (gray: native; red: \rsipde{}).
    Right: learning curves (black: native; red: \rsipde{}). Both halves have Opus~5.5 above GPT-5.5,
    with Full, Noisy and Hard-OOD columns. Labels give final counts and
    RSI-minus-native case gains. Axes are locally zoomed; cropped bar baselines are marked.
    The native Opus Full curve whisker shows the observed 187--198 range over three runs. Curves preserve regressions
    and stopping rounds. \cref{tab:main,tab:settings} give all models and baselines.}
  \label{fig:teaser}
\end{figure}

\section{Introduction}
\label{sec:intro}

Scientific observations rarely arrive with a fully specified governing equation. Sparse field measurements
and known forcing or boundary conditions may leave both the physical law and its coefficients uncertain.
Noise and fractional dynamics make this identification harder. An accurate discretization of the wrong
equation is insufficient, and a correct family with inaccurate coefficients can still produce a poor solution.

Coding agents can explore these possibilities through numerical experiments, but executable code alone is
not evidence of scientific accuracy. Existing datasets support learning PDE solution operators and discovering
equations from data \citep{takamoto2022pdebench,gupta2022pdearena,hao2024pinnacle,ohana2024well,rudy2017pdefind}.
CodePDE studies solver synthesis for specified equations \citep{li2025codepde}, and PDEAgent-Bench provides
645 specified-PDE solver-generation instances \citep{hang2026pdeagentbench}. We ask whether an agent can
instead identify an unknown PDE or fractional PDE (FPDE), estimate its parameters and construct an accurate
solver within a limited development budget.

We introduce \bench{}, a collection of \textbf{1,001 released PDE and FPDE cases across 42 templates} for this
identify-then-solve task. To our knowledge, it is the largest joint PDE/FPDE benchmark for coding agents tasked
with equation-family identification, parameter estimation and numerical solver construction. This scope concerns
agent problem instances; it does not compare the number of simulated trajectories in general-purpose PDE
learning datasets. The agent receives a finite equation library, 256 spatial--temporal samples and known
forcing, initial and boundary functions, while the true family and coefficients remain hidden for a new case.
This is a controlled analogue of scientific system identification, with classical, nonlinear, multiphysics
and fractional dynamics in a common executable interface.

Three complementary settings expose distinct sources of uncertainty. \textbf{Full} tests all 42 templates
with noise-free observations. \textbf{Noisy} adds Gaussian perturbations at 50\% of the clean-signal RMS.
\textbf{Hard-OOD} combines unseen templates and coefficients shifted beyond development ranges. Together,
they probe instance generalization, observation robustness and transfer to unfamiliar physics or parameters.
Their held-out counts are distinct from the 1,001-case release.

The \textbf{\rsipde{} harness} organizes exploration through residual guidance and reference-based validation.
Inside a sandbox, the agent substitutes a reconstructed or computed field into candidate equations, fits
coefficients and retains up to five low-residual families. Forward solves and public relative $L^2$ feedback
then distinguish hypotheses and refine coefficients and discretizations. A gate retains improvements to the
solver, its instruction policy and memory, while the foundation model and executable supervisor stay fixed.
Residuals narrow the family search; relative $L^2$ tests assess the resulting solution. Hidden references
never guide online equation choice.

We compare single-run programs, native harnesses (Claude Code or Codex), and \rsipde{} using Opus~5.5,
GPT-5.5, GPT-6.1 and Sonnet~5.5. Across the four models, \rsipde{} increases mean pass rate over single-run
baselines by 181.9\%, 140.2\% and 106.5\% on Full, Noisy and Hard-OOD, and over native harnesses by 3.0\%,
19.6\% and 5.4\%, respectively. Recorded model calls sum to 1,884 rather than 3,547 across the 12 comparisons,
a 46.9\% reduction, although calls and dollar costs are not lower in every comparison. \cref{fig:teaser}
shows the final counts and learning curves for Opus~5.5 and GPT-5.5. These are descriptive results from the
reported runs; replication, adaptive benchmark reuse and cost provenance are discussed in \cref{sec:limits}.

Our contributions are:
\begin{itemize}
 \item \textbf{A large joint PDE/FPDE benchmark:} 1,001 released cases spanning 42 equation templates, with
       analytic or high-accuracy numerical references and quantitative solution checks.
 \item \textbf{Three diagnostic evaluation settings:} Full, Noisy and Hard-OOD separately test noise-free
       accuracy, robustness to 50\% Gaussian sample noise, and unseen templates with shifted coefficients.
 \item \textbf{An unknown-family, unknown-parameter task:} agents must explore a finite physical hypothesis
       space, estimate coefficients and return an executable solver from sparse observations and known inputs.
 \item \textbf{A residual-guided improvement harness:} equation residuals prioritize a five-candidate shortlist,
       while public relative $L^2$ feedback and validation retain improvements to equation selection,
       coefficient estimation, numerical solution, policy and memory.
\end{itemize}

The remainder introduces the task and evaluation interface (\cref{sec:bench}), dataset construction
(\cref{sec:construction}), the harness (\cref{sec:harness}) and experiments (\cref{sec:experiments}).
Related work and implementation details appear in the appendix. Code, data and the result website are linked
on the first page.

\begin{figure}[t]
  \centering
  \includegraphics[width=\textwidth]{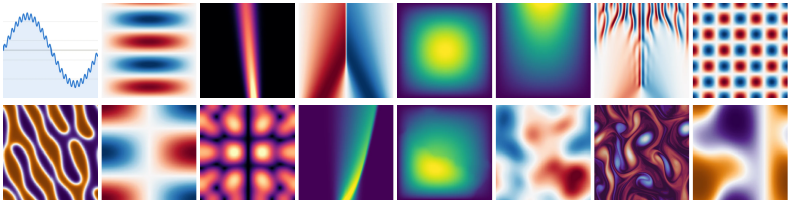}
  \caption{Examples of reference fields from public training cases. The benchmark includes diffusion and
    transport, fractional dynamics, wave propagation, fluid flow, phase-field systems and electromagnetics.
    The gallery is illustrative; \cref{app:templates} gives the complete 42-template catalog.}
  \label{fig:gallery}
\end{figure}

\section{Benchmark Overview}
\label{sec:bench}

\subsection{Task scope: unknown equation and unknown coefficients}
\label{sec:task}

A case is generated from
\begin{equation}
 \mathcal N_{k,\theta}[u](x,t)=f(x,t),\qquad (x,t)\in\Omega\times(0,T],
 \label{eq:task}
\end{equation}
with the corresponding steady or ordinary-differential specialization. The hidden index $k$ selects a
PDE/FPDE from the supplied class library $\mathcal K$, and $\theta$ denotes its unknown physical coefficients,
including fractional-order parameters where applicable. The agent must infer both before choosing and
implementing a suitable numerical method. Template names in the library define the allowed hypothesis space;
they do not reveal the class of an individual hidden case.

This formulation captures a difficult part of practical scientific modeling: observations can fit several
candidate laws, and model mismatch, parameter error and discretization error can look similar. It remains a
controlled finite-library task, rather than unrestricted discovery of an arbitrary physical law. The benchmark
makes these uncertainties explicit so that exploration and numerical repair can be evaluated together.

\subsection{Agent-facing inputs and executable interface}
\label{sec:interface}

The public case view contains the domain, known callable forcing, initial and boundary functions,
evaluation-grid specifications, the equation library, and 256 spatial--temporal samples,
\begin{equation}
 \mathcal D_u=\{(z_i,y_i)\}_{i=1}^{256},\qquad y_i=u^\star(z_i)+\epsilon_i.
 \label{eq:observations}
\end{equation}
Here $z_i=(x_i,t_i)$ denotes a sample location, specialized to spatial-only or time-only coordinates where
appropriate. Full sets $\epsilon_i=0$; the other settings perturb observed values as described in
\cref{sec:splits}. 

An entrant writes \texttt{solve(problem)}. Its program identifies a candidate class and coefficients and
returns a numerical solution on its chosen grid, or at prescribed points for the Elliptic equation problem.
The evaluator maps the returned field to its evaluation grid. Class and coefficient estimates are diagnostic
outputs; a correct class name without an accurate field does not pass. The entrant can choose classical,
learned or hybrid numerical methods within the sandbox and resource contract.

Training diagnostics can reveal class and coefficient labels. Validation returns numerical diagnostics but
not the hidden class or coefficients. Hidden test labels, reference arrays and scores are inaccessible during
agent execution. A retained program must therefore identify each new test case from its public inputs,
without selecting an equation using that case's hidden reference error.

\subsection{Accuracy, executability and reference solutions}
\label{sec:reference}

The reference field $u^\star$ defines the numerical scoring target. For analytic cases, the evaluator computes
its values directly at evaluation nodes. Other references are produced with refined meshes, suitable
higher-order discretizations or time integration, and strict solver tolerances. Fine/coarse calculations
mapped to a common grid check self-convergence. Reference generation and thresholds are fixed independently
of an entrant's solver; \cref{app:references} gives representative settings and their limitations.

Let $\mathcal I\hat u$ be the submitted solution interpolated to the reference grid. For time-dependent problems,
\begin{equation}
 E_{\mathrm{st}}=
 \frac{\|\mathcal I\hat u-u^\star\|_{L^2(\Omega\times[0,T])}}
      {\|u^\star\|_{L^2(\Omega\times[0,T])}},
 \qquad
 E_T=
 \frac{\|\mathcal I\hat u(\cdot,T)-u^\star(\cdot,T)\|_{L^2(\Omega)}}
      {\|u^\star(\cdot,T)\|_{L^2(\Omega)}} .
 \label{eq:l2}
\end{equation}
The norms use the evaluator's quadrature weights; steady and point-cloud cases use their corresponding norms.
The submitted program passes only if both applicable errors are at most $\tol=2\times10^{-3}$, initial and
boundary conditions satisfy the prescribed $10^{-10}$ check, the returned arrays meet the contract, and the
solve meets runtime and memory limits. The Elliptic equation problem has a separately fixed
$5\times10^{-2}$ tolerance because its reference discretization is less accurate. The model does not receive
the hidden reference arrays. Public training and validation errors are computed by the supervisor.

Pass count and pass rate summarize end-to-end success. Relative $L^2$ diagnostics distinguish numerical
accuracy from execution or contract failure, while model calls, tokens, validation submissions and dollar
cost characterize development effort. Class labels and parameter estimates help diagnose failures but are
not substitutes for the solution-error criterion.

\subsection{Budgeted development and frozen testing}
\label{sec:protocol}

A complete experiment combines three development sub-runs organized by solver-library group. Each receives
the allowance in \cref{tab:budget}; groups may run in parallel. Across three groups the declared ceilings
are 1,200 model calls, 120\,M prompt tokens, 4.5\,M output tokens and 90 validation submissions. The two-hour
allowance applies to each group. Aggregate usage can therefore exceed a single group's 400-call limit.

\begin{table}[t]
  \caption{Declared budget per development sub-run. All aggregate resource figures sum the three sub-runs
    before averaging repeated runs. Final hidden evaluation is outside the development budget.}
  \label{tab:budget}
  \centering
  \begin{small}
  \begin{tabular}{lr}
    \toprule
    Resource & Cap\\
    \midrule
    Development wall-clock & 2\,h\\
    Model requests & 400\\
    Prompt tokens, including cache reads/writes & 40\,M\\
    Output tokens & 1.5\,M\\
    Validation submissions & 30\\
    Solver resources & 1 CPU core + 1 H100 GPU\\
    \bottomrule
  \end{tabular}
  \end{small}
\end{table}
A proxy records model identity and usage. Model-written code runs in a separate sandbox with bounded imports,
filesystem access, network access, runtime and memory. Agents can request training experiments within the
development budget; a validation submission is counted when evaluation starts. The benchmark-native wrapper
retains accepted programs using the reference-error gate. \cref{sec:gate} describes the \rsipde{} retention
mechanism, including its public-pass rule.

After development, an independent evaluator scores the frozen retained program on hidden cases. For a combined
learning curve at round $r$, each group contributes its retained program after $r$ validation submissions,
or its final program if it stopped earlier. Thus a round is a submission index, not a common elapsed-time
budget, and summed group submissions differ from the final index of a combined curve. Hidden checkpoint
scores are assembled for reporting and do not return to the running agent.

\section{Dataset Construction}
\label{sec:construction}

\subsection{Case inventory and equation coverage}
\label{sec:coverage}

\bench{} releases 1,001 cases and covers 42 templates. A template specifies an equation and a construction
of coefficients, forcing, initial/boundary data and reference solution; an individual case instantiates that
construction. Multiple templates may share one equation family, so 42 templates does not mean 42 unrelated
physical laws. The catalog contains 20 classical and fractional templates, 10 nonlinear or multiphysics
templates, and 12 data-driven or additional-dimensional templates (\cref{app:templates}).

Classical coverage includes Poisson, heat, convection--diffusion, Burgers and wave equations. Fractional
coverage includes a Caputo ODE, time-fractional diffusion, multi-term fractional dynamics and orders varying
in time or space--time. These problems test whether identification routines can distinguish local dynamics
from memory-dependent operators and estimate their coefficients or orders. Nonlinear and multiphysics
coverage includes Kuramoto--Sivashinsky, Navier--Stokes, Cahn--Hilliard, Maxwell and time-harmonic
 electromagnetics. Additional templates include Darcy, Helmholtz, Allen--Cahn, two-dimensional Burgers and
advection--diffusion--reaction equations.

Spatial dimensions are one or two, with the fractional ODE as the zero-spatial-dimensional specialization.
The collection mixes manufactured solutions, freely evolving initial conditions, sharp layers, multiscale
fields and nonlinear dynamics. Five templates reuse input fields from DiffusionPDE
\citep{huang2024diffusionpde}; their reference solutions are recomputed for this benchmark. These fields add
variation in coefficients or inputs without making the full collection a dataset of experimental measurements.
\cref{fig:gallery} illustrates public training fields.

\subsection{Full, Noisy and Hard-OOD settings}
\label{sec:splits}

The three settings probe different sources of uncertainty. They are not an exhaustive partition of the
1,001 released cases, and their pass rates describe different case distributions. Full draws a balanced
noise-free evaluation across the catalog, Noisy selects a diagnostic portion of the public release, and
Hard-OOD uses a separate sealed draw. Their training, validation and hidden-test allocations are given in
\cref{tab:splits}.

\begin{table}[t]
  \caption{The three current evaluation settings. Noisy contains 200 selected cases in total, of which 160 are
    test cases. Hard-OOD separates 188 cases from unseen templates and 90 cases with shifted coefficients.}
  \label{tab:splits}
  \centering
  \begin{small}
  \begin{tabular}{lrrrrl}
    \toprule
    Setting & Templates & Train & Validation & Test & Observation noise\\
    \midrule
    Full & 42 & 42 & 42 & 210 & None\\
    Noisy & 24 & 22 & 18 & 160 & 50\% RMS\\
    Hard-OOD & 39 & 20 & 20 & 278 & 1\% RMS\\
    \bottomrule
  \end{tabular}
  \end{small}
\end{table}
\paragraph{Full: noise-free instance generalization.}
Full uses the \texttt{full0} sampler, with no observation noise. Each of the 42 templates contributes one
training, one validation and five hidden-test cases: 42/42/210 in total. Public seeds determine training
and validation instances, while a separate secret determines hidden tests. It measures generalization to
new instances from represented template distributions.

\paragraph{Noisy: severe observation corruption.}
Noisy selects 200 cases found easiest by fixed calibration probes and adds Gaussian perturbations with
standard deviation equal to 50\% of the clean-signal RMS. The selected cases cover 24 templates and yield
22 training, 18 validation and 160 test instances. A template with at least three selected cases contributes
to all splits; two-case templates contribute training and test instances, and one-case templates contribute
test instances. Selection helps distinguish observation robustness from baseline numerical difficulty.
Because these cases come from a regenerable public release, this setting is a diagnostic test rather than
a fresh secret holdout.

\paragraph{Hard-OOD: unseen templates and shifted coefficients.}
A sealed draw over 39 templates provides 20 training and 20 validation cases from 20 seen templates.
Its 278 hidden cases combine 188 instances from 19 templates absent from both development splits and
90 instances with coefficients in the opposite half of the range used in development. The latter come
from the seen templates with an identifiable coefficient; the two seen templates without such a coefficient
contribute no shifted cases. Observation noise is 1\% RMS and its level is hidden. The equation library is
still supplied, so unseen means absent from training and validation, not absent from the allowed library.
The task text discloses both forms of shift. A problem-specific feasibility filter conditions the draw,
limiting the population to which its results generalize.

\subsection{Reference validation and reproducibility}
\label{sec:dataset-validation}

Case generation keeps stable identifiers, template lineage and separate public/evaluator views. Analytic
references are sampled directly on the scoring grid; numerical references use family-appropriate refinement
and convergence checks (\cref{app:references}). Fixed solver probes inform calibration and hardening, but do
not change the acceptance tolerance for an individual entrant. The same frozen evaluator checks all reported
programs for a setting.

The release, split construction, numerical references and frozen result snapshot support separate inspection
of task generation, solver behavior and reported aggregates. Neither the 1,001-case inventory nor the held-out
pass count measures unrestricted scientific coverage: dimensions, templates, admissible parameters and
feasibility conditioning define the benchmark's scope.

\section{RSI-PDE Harness}
\label{sec:harness}

\subsection{Fixed supervisor and evolving solver, policy and memory}
\label{sec:state}

\begin{figure}[t]
 \centering
 \includegraphics[width=\textwidth]{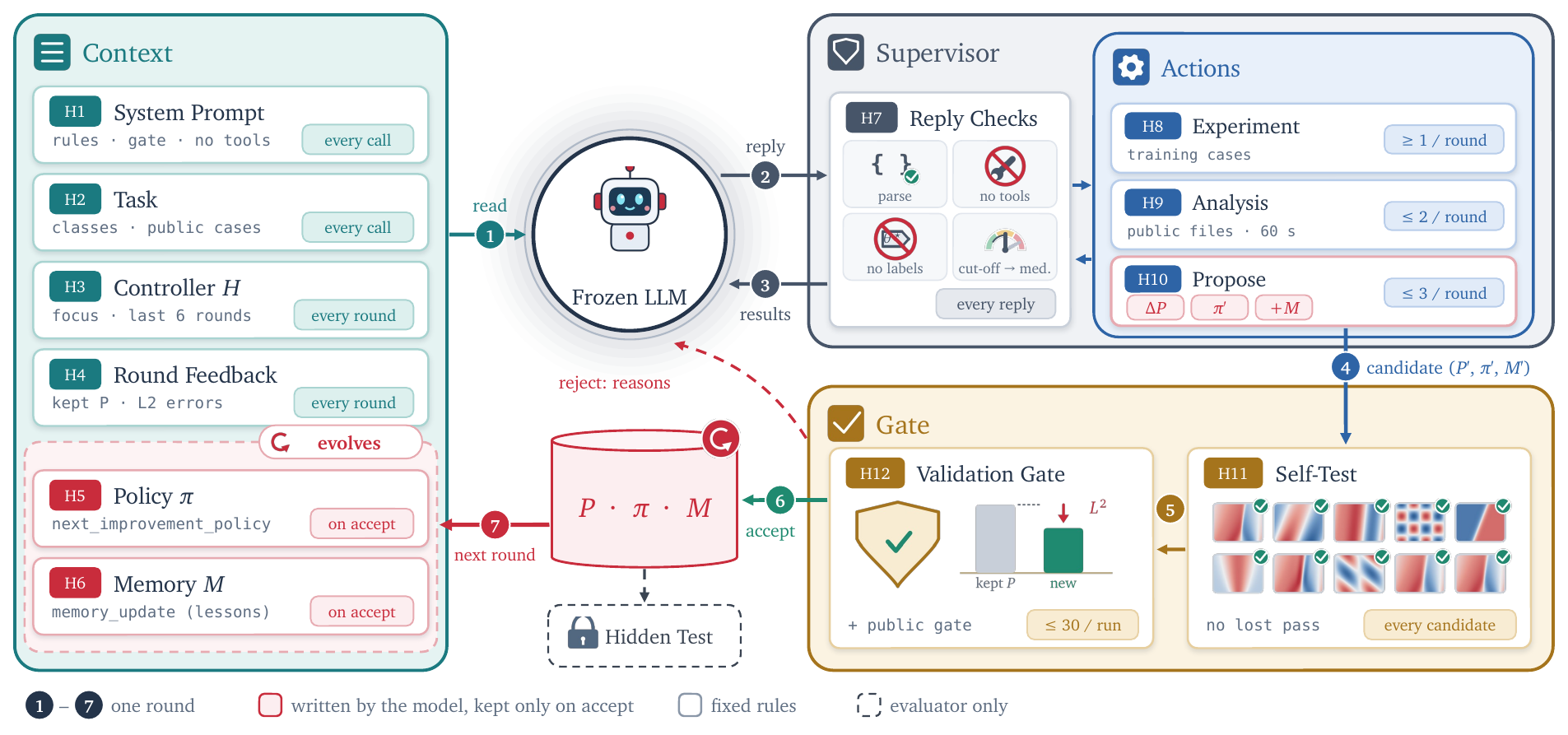}
 \caption{\textbf{RSI-PDE harness.} The frozen LLM reads the context (H1--H6), and its reply passes
 the supervisor checks (H7) before becoming a training experiment (H8), a public-data analysis (H9) or a
 proposal (H10). A proposal updates solver $P$, instruction policy $\pi$ and memory $M$. The candidate
 passes the training self-test (H11) and the validation or public-pass gate (H12); the text describes the
 bounded training-regression waiver.
 Accepted state returns in the next round's context; rejected proposals return diagnostic feedback.
 The supervisor and model weights remain fixed. Only the independent evaluator scores the frozen solver
 on hidden tests.}
 \label{fig:harness}
\end{figure}

\begin{figure}[t]
 \centering
 \includegraphics[width=\textwidth]{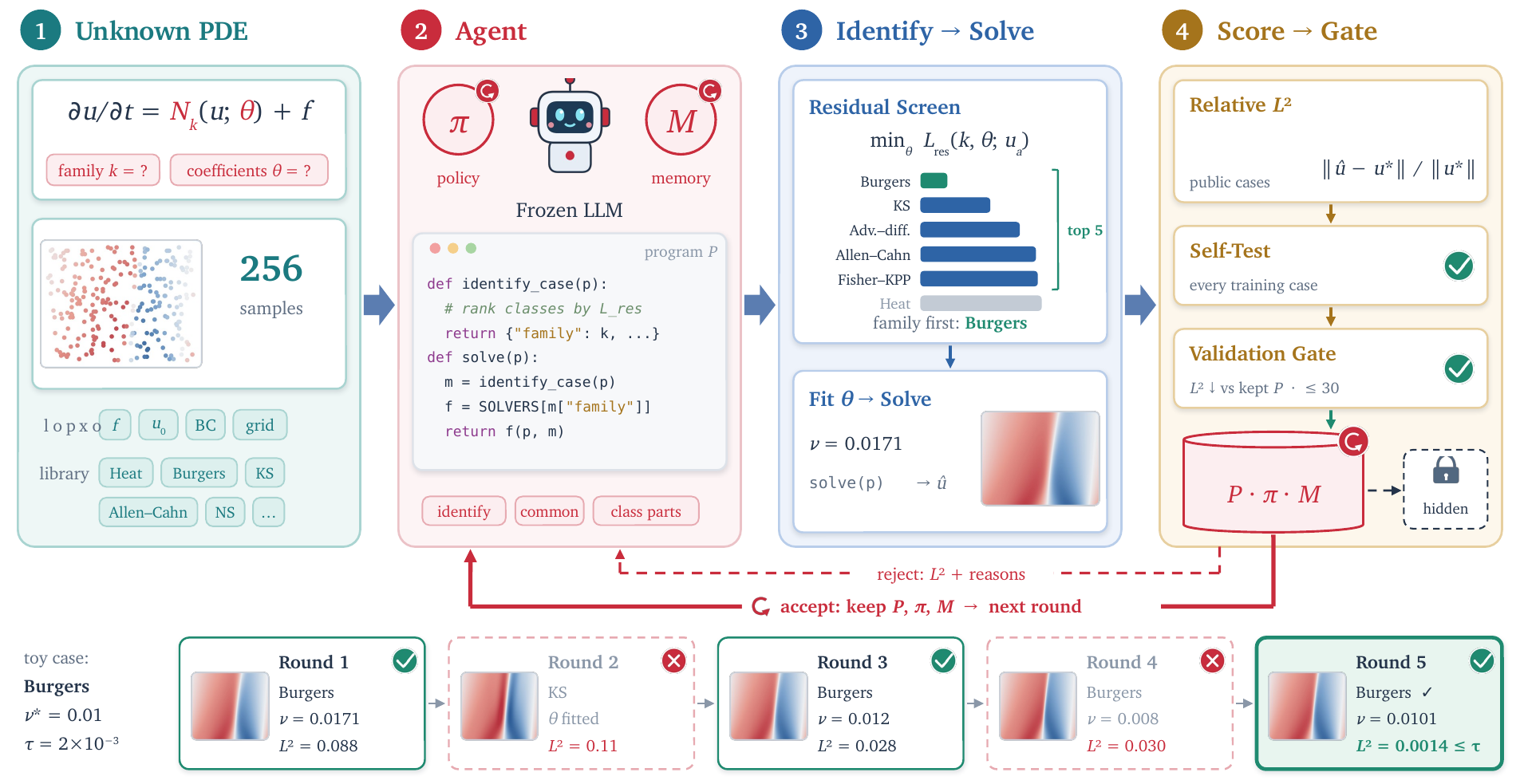}
 \caption{\textbf{The residual-guided RSI-PDE workflow.} (1) The agent starts without the equation: the
 family $k$ and the coefficients $\theta$ are unknown, and it receives 256 samples, the known forcing, initial
 and boundary data, and the class library. (2) A frozen LLM, steered by its self-written policy $\pi$ and
 memory $M$, writes the program $P$, whose router identifies the case and dispatches it to a class solver.
 (3) In the sandbox, residual ranking by $\min_\theta\mathcal L_{\mathrm{res}}(k,\theta;u_a)$ ranks candidate
 families and keeps the five best; forward solves and public relative $L^2$ checks then refine the equation
 choice and its coefficients. (4) Public relative
 $L^2$ errors, the self-test and the validation gate decide what is kept. Accepted $(P,\pi,M)$ carry into the
 next round, so the policy and memory evolve with the solver; a rejection returns its $L^2$ errors and
 reasons. The bottom row follows a toy viscous Burgers case ($\nu^\star=0.01$): the residual fit on a smoothed
 field gives $\nu=0.0171$, and gated $L^2$ feedback refines it to $\nu=0.0101$, below $\tau$. The later
 coefficient values are illustrative; the residuals, fields and errors are computed for this toy case, not
 taken from benchmark runs.}
 \label{fig:workflow}
\end{figure}

The \rsipde{} harness structures an agent's numerical exploration while keeping its foundation model and
executable supervisor fixed. Its accepted state is $(P,\pi,M)$: $P$ is a modular solver with an identification
router, equation-class modules and shared helpers; $\pi$ is a model-written instruction policy for subsequent
investigation; and $M$ stores reusable numerical lessons. \cref{fig:harness} shows how that state moves through
one round. \cref{fig:workflow} separately follows the physical identification and refinement loop.

Each round supplies the active program, policy, memory, recent history, public diagnostics and remaining
resources. The agent can analyze public files, experiment on training cases and propose solver/policy/memory
updates. The supervisor requires a training experiment before a proposal and bounds the round to at most two
analyses and three proposal attempts. Reply parsing, sandbox execution, label-copy checks, no-op filtering
and training self-tests constrain these actions before a validation submission is spent. Numerical analysis
runs inside the solver sandbox; it does not grant the model unrestricted host execution.

\subsection{Residual guidance for PDE and FPDE family search}
\label{sec:residual-first}

The agent begins with an available field $u_a$, obtained from its current solver or a regularized
reconstruction of public observations. It substitutes this field into candidate equations inside the sandbox
and fits their unknown coefficients. For candidate class $k$, write the equation as signed terms,
including the forcing:
\begin{equation}
 \mathcal R_{k,\theta}[u_a]=\sum_j T_{k,j}[u_a;\theta],\qquad
 r_{\mathrm{id}}(k,\theta)=
 \frac{\|\mathcal R_{k,\theta}[u_a]\|_{\mathcal Q}}
      {\sum_j\|T_{k,j}[u_a;\theta]\|_{\mathcal Q}+\varepsilon},
 \quad \mathcal L_{\mathrm{res}}=r_{\mathrm{id}}^2 .
 \label{eq:id-residual}
\end{equation}
Here $\mathcal Q$ denotes the usable collocation grid or weak-test set, and $\varepsilon$ protects against a
zero normalizer. Coefficients are fitted within the candidate's admissible ranges. Class-specific objectives
may also include sample-fit and boundary terms. The candidate law and public inputs define this calculation;
neither the hidden true equation nor a hidden reference field is required.

Residual ranking prioritizes at most five hypotheses:
\begin{equation}
 \mathcal C_5=\operatorname{TopK}_{k\in\mathcal K}
       \left[-\min_{\theta}\mathcal L_{\mathrm{res}}(k,\theta)\right],
 \qquad K=\min(5,|\mathcal K|).
 \label{eq:top5}
\end{equation}
This screening focuses subsequent forward-solver experiments on a small set of plausible PDE/FPDE families.
It avoids committing to one equation solely because it fits sparse observations. For noisy data, numerical
differentiation can amplify perturbations; the agent can use smoothed reconstructions, solver-generated
fields, spectral representations or weak residuals suited to the candidate operator. Term normalization and
known initial/boundary data help distinguish physical consistency from an artificially favorable term scale.

The purpose of residual guidance is to reduce unstructured family search. The aggregate experiments do not
isolate its wall-clock benefit or top-five recall, and a low residual does not guarantee that the correct law
survives the shortlist. Residual computation remains an agent-written numerical procedure, rather than a
supervisor oracle revealing the true equation.

\subsection{Relative \texorpdfstring{$L^2$}{L2} feedback and coefficient refinement}
\label{sec:refinement}

For each shortlisted family, the agent implements or revises a forward solver, estimates coefficients and
checks its predicted field. Residual and sample-fit evidence suggest candidate parameters; public training
relative $L^2$ errors show whether the resulting solution is accurate. Repeated experiments can distinguish
a wrong family from a poorly fitted coefficient or an under-resolved numerical method. Known forcing,
initial and boundary functions provide additional constraints on an otherwise ambiguous inverse problem.

The combination of residual and relative $L^2$ information is staged: residual loss prioritizes scientific
hypotheses, then reference-based tests guide equation choice, coefficient estimation and discretization
repair on public cases. The validation gate evaluates whether a proposed reusable program improves. There
is no arbitrary weighted residual-plus-$L^2$ threshold in the pass criterion. The desired outcome is a
coefficient estimator and solver that recover accurate solutions on new instances; the solution pass rate
alone does not establish exact coefficient recovery when different parameters are observationally similar.

At hidden inference time, the frozen router must choose from public inputs, residuals and sample-fit evidence.
It cannot request the test reference or its relative $L^2$ error. Public development errors improve the
identification procedure; they do not provide an online oracle for selecting the hidden test equation.

\subsection{Validation, retention and feedback}
\label{sec:gate}

The standard reference-error gate requires a greater than 0.5\% decrease in macro relative $L^2$ error,
no class-average increase above 5\%, no loss of class passes, no per-case space--time or final-time error
increase above 25\%, and valid execution and initial/boundary data. Class-local changes can be gated and
merged independently; shared router/helper changes are evaluated at program level.

The harness also uses a public-pass rule over training and validation cases when the standard gate rejects
a candidate. That rule requires at least one gained pass, no lost pass, no newly broken execution or
constraint/resource failure, and at most a 25\% error increase for cases passing in both programs, subject
to the implementation's numerical floor. The all-training self-test normally preserves existing training passes. After two consecutive rounds
blocked by that guard, it can waive the training-regression rule for one round; the validation/public-pass
decision still applies. Thus an update can address a remaining public failure without being retained solely for reducing
errors on already accurate cases.

Accepted solver edits carry their corresponding policy and memory updates into the next round. Rejected
candidates return their numerical errors and rejection reasons, while remaining in the audit history.
\cref{alg:rsipde} summarizes this process. The solver, policy and memory evolve; the displayed experiments
do not promote self-modified executable supervisors. Establishing improvement of the controller itself
would require equal-budget paired continuations from the same starting solver and memory.

\begin{algorithm}[t]
 \caption{Residual-guided solver development in \rsipde{}}
 \label{alg:rsipde}
 \begin{algorithmic}
  \STATE \textbf{Input:} fixed model, public library and data, development budget
  \STATE Initialize solver $P$, instruction policy $\pi$ and memory $M$
  \WHILE{development resources and validation submissions remain}
   \STATE Obtain public diagnostics and the current field approximation $u_a$
   \STATE In the sandbox, fit candidate equations and evaluate $\mathcal L_{\mathrm{res}}$
   \STATE Form or refresh the shortlist $\mathcal C_5$ of at most five equations
   \STATE Allocate experiments to shortlisted solvers; compare residuals and training $L^2$ errors
   \STATE Refine the equation, coefficients and numerical method; propose $(P',\pi',M')$
   \STATE Apply static checks, training self-tests and proposal filters
   \IF{the candidate is eligible for validation}
    \STATE Evaluate reference-based validation errors and apply the acceptance gate
    \IF{accepted}
     \STATE Merge the accepted program edits and corresponding policy/memory updates
    \ENDIF
   \ENDIF
   \STATE Record diagnostics, proposal lineage and resource use
  \ENDWHILE
  \STATE Freeze the retained solver; evaluate test accuracy and post-hoc physical residuals
 \end{algorithmic}
\end{algorithm}

\section{Experiments}
\label{sec:experiments}

\subsection{Experimental setup and baselines}
\label{sec:provenance}

We compare four foundation models---Opus~5.5, GPT-5.5, GPT-6.1 and Sonnet~5.5---under three modes:
\begin{itemize}
 \item \textbf{Single run:} one model response per development group, frozen without iterative repair.
 \item \textbf{Native harness:} Claude Code for the Claude models and Codex for the GPT models, with
       execution and repair under the benchmark wrapper and declared development limits.
 \item \textbf{\rsipde{} harness:} structured numerical experiments and gated solver, policy and memory
       updates as described in \cref{sec:harness}.
\end{itemize}
Model weights stay fixed within each model-specific comparison. Native and \rsipde{} modes share the declared
resource ceilings and independent numerical evaluator, but their tools and prompting differ; this is a
system comparison rather than an ablation of one component. We report pass counts, relative $L^2$
diagnostics, learning curves and resource use separately.

The numerical record is the frozen public website snapshot dated November 7, 2026, including corrected
GPT-6.1 Noisy identities. Most harness cells contain one run; native Opus~5.5 on Full averages three runs,
with 187--198 passes. Single-run means use the available response samples for each model and setting.
The Opus single-run baseline uses 128k output tokens. Where an answer was re-requested at the model's maximum
output length, it replaces the answer truncated by a shorter limit; incomplete one-response programs score
zero. Unequal replication and output allowances should be considered when interpreting the comparisons.

Author-reported single-run totals are GPT-6.1 Hard-OOD 135/278 and Sonnet~5.5 Full 59/210, Noisy 36/160 and
Hard-OOD 104/278. Their missing validation/resource breakdowns are not inferred. Full native GPT-6.1 and
Sonnet~5.5 headline totals are 189 and 183, with curves supplied as rounded percentages. Their earlier
per-template, residual and resource measurements remain explicitly labeled diagnostic records, rather than
being reconstructed to match these totals. GPT-6.1 Noisy uses the corrected attribution throughout:
Codex 90/160 and \rsipde{} 97/160. The snapshot preserves raw record identifiers and source provenance;
\cref{app:implementation} describes reproduction.

\paragraph{Aggregate improvement.}
Across the four models, mean pass rates for single run/native/\rsipde{} are 28.7/78.6/81.0\% on Full,
26.5/53.3/63.8\% on Noisy and 37.2/72.9/76.9\% on Hard-OOD. Relative gains are computed as
$100(\bar p_{\mathrm{RSI}}-\bar p_{\mathrm{baseline}})/\bar p_{\mathrm{baseline}}$, using the mean of the four
model pass rates within each setting. These yield 181.9/140.2/106.5\% gains over single run and
3.0/19.6/5.4\% over native harnesses. The corresponding absolute native-to-\rsipde{} gains are
2.4/10.5/4.0 percentage points. Relative gains and percentage-point gains therefore have different scales.

\subsection{Full benchmark}
\label{sec:main}

\begin{table}[t]
 \caption{Full results. Passes are out of 210; gain over the single-run baseline and validation-minus-test
 gap are percentage points.
 Bold marks the higher displayed harness count for a model.
 $\dagger$: author-updated native total with earlier-run resource/diagnostic records.
 $\ast$: retrospective stop-adjusted RSI cost, not the cost of the full-budget score shown.}
 \label{tab:main}
 \centering
 \resizebox{\textwidth}{!}{\input{tables/main_full.tex}}
\end{table}

\cref{tab:main} gives all four models. \rsipde{} reaches 196/210 (93.3\%) with Opus~5.5, 104/210 (49.5\%) with
GPT-5.5, 193/210 (91.9\%) with GPT-6.1 and 187/210 (89.0\%) with Sonnet~5.5. The respective native counts are
193, 95, 189 and 183. The observed margins are therefore 3, 9, 4 and 4 cases, or approximately 1.4, 4.3, 1.9
and 1.9 percentage points. These are descriptive differences, not significance results; the three-case
Opus difference is smaller than the native harness's observed 11-case range.

Across all four models, iterative harnesses add approximately 57--130 passes relative to the available
single-run baselines. For Sonnet~5.5, the native harness improves from 59 to 183 passes and \rsipde{} to 187. This supports the practical value of execution and repair under
numerical feedback. It does not isolate a particular residual, memory or acceptance rule: the system comparison changes
multiple aspects of exploration and feedback together.

\begin{figure}[t]
 \centering
 \includegraphics[width=0.9\textwidth]{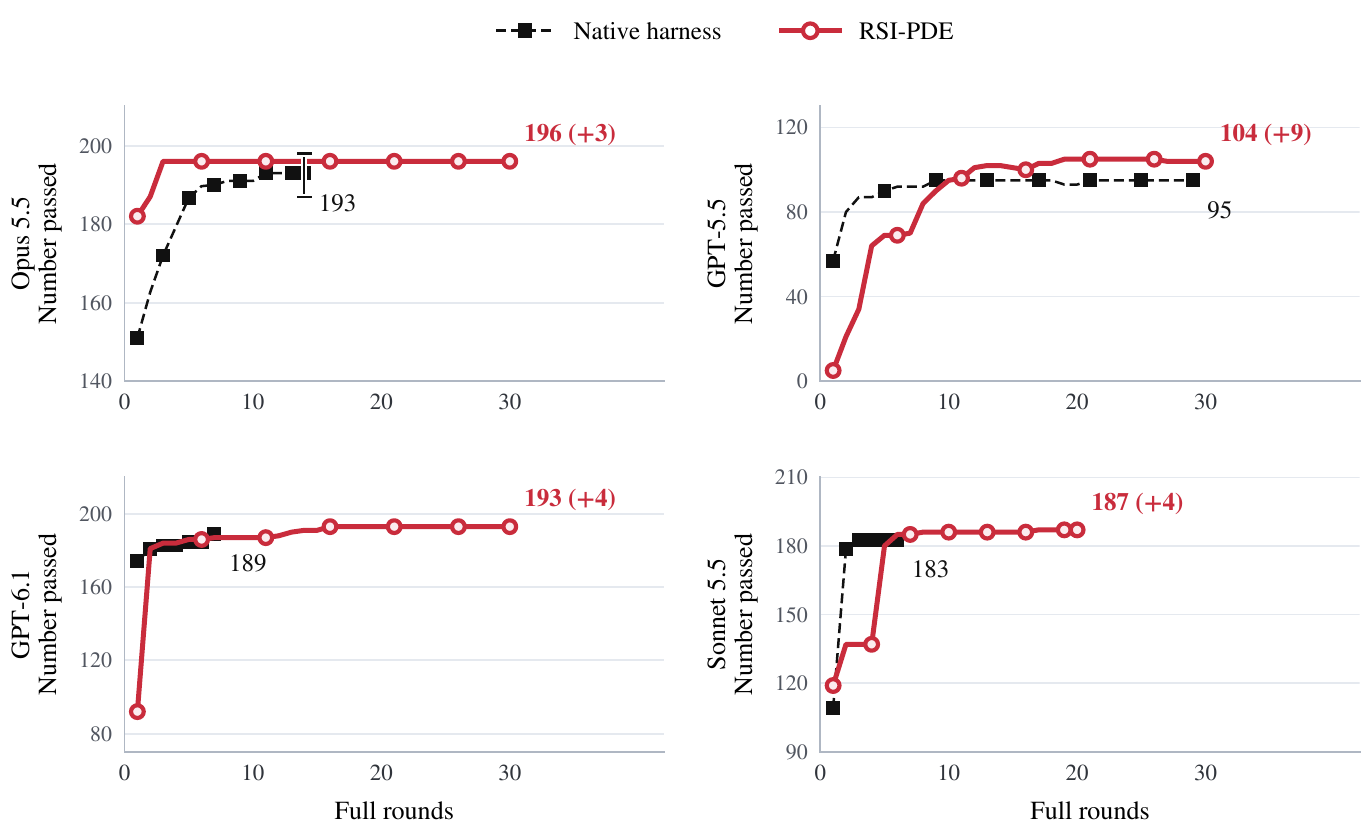}
 \caption{Full passed-case trajectories for all four models.
 Black dashed squares: native harness; red circles: \rsipde{}. Local vertical ranges show recorded points
 from round one. Final labels give headline counts; red labels also give RSI-minus-native gains in cases.
 The native Opus~5.5 whisker shows the observed 187--198 range over three runs. GPT-6.1 and Sonnet~5.5
 native count trajectories are approximate, converted from the author's rounded percentages; the Sonnet
 endpoint therefore need not equal 183 before rounding. A round index is not a common wall-clock budget.}
 \label{fig:curves}
\end{figure}

The curves in \cref{fig:curves} show strong early gains and later plateaus for several models.
They are most useful as records of retained solver quality: repeated public submissions and unchanged
programs can leave long flat segments. Since sub-runs can stop at different times, the horizontal axis does
not establish that one system reached an accuracy level faster in seconds.

\subsection{Severe noise and out-of-distribution generalization}
\label{sec:ood}

\begin{table}[t]
 \caption{Noisy and Hard-OOD results for all four models. Bold marks the higher harness count, including ties.
 Unseen and shifted counts partition Hard-OOD. $\ddagger$: corrected GPT-6.1 Noisy attribution; the mark
 applies to the Noisy identity and its corresponding calls. Missing single-run group breakdowns remain
 unavailable rather than inferred.}
 \label{tab:settings}
 \centering
 \resizebox{\textwidth}{!}{\input{tables/settings.tex}}
\end{table}

On Noisy, \rsipde{} passes 120/160 with Opus~5.5, 93/160 with GPT-5.5, 97/160 with GPT-6.1 and 98/160 with
Sonnet~5.5, compared with 76, 83, 90 and 92 for the native harnesses. The largest observed margin is 44 cases
with Opus~5.5. The corrected GPT-6.1 comparison favors \rsipde{} by seven cases. These differences suggest
that investigation and coefficient estimation are particularly consequential under heavy observation noise,
but a residual-shortlist ablation is needed to attribute that effect to the proposed identification procedure.

Hard-OOD yields 254/278 versus 245 with Opus~5.5, 102 versus 81 with GPT-5.5, 261 versus 261 with GPT-6.1,
and 238 versus 224 with Sonnet~5.5. The reported GPT-6.1 single-run baseline passes 135/278, so both its native and RSI harnesses add 126 cases
(45.3 percentage points), while remaining tied with each other. On the 90 shifted-coefficient cases,
\rsipde{} passes 89, 45, 88 and 90,
while the native harnesses pass 84, 37, 86 and 90. Thus it is no worse on the displayed shifted comparisons,
with strict gains for three models and a tie for Sonnet~5.5. The unseen-template comparison is less uniform:
GPT-6.1 favors Codex by two cases (175 versus 173), despite the equal aggregate result.

These observations motivate keeping equation selection, coefficient identification and forward discretization
as separate diagnostics. A transferable identification routine must fit a new instance's parameters, while
a broad solver library must still route templates absent from development. Hard-OOD is conditioned by its
construction and each cell has one run, so the current results do not establish universal transfer behavior.

\subsection{Model calls and cost}
\label{sec:cost}

\begin{figure}[t]
 \centering
 \includegraphics[width=0.95\textwidth]{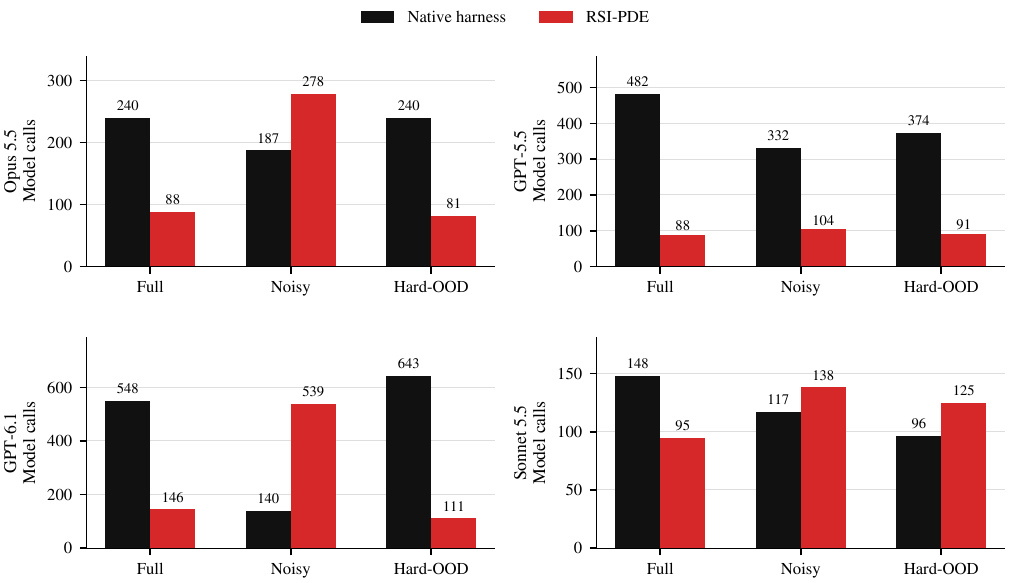}
 \caption{Recorded model calls by model and setting. Black: native harness; red: \rsipde{}. Calls are summed
 across development groups and are not replaced by stop-rule replay counts. The corresponding dollar
 figures have a different provenance, as explained in the text and \cref{tab:costs}.}
 \label{fig:calls}
\end{figure}

Across the 12 model--setting pairs, recorded calls sum to 3,547 for native harnesses and 1,884 for
\rsipde{}, a 46.9\% reduction. On Full, the native-to-RSI call ratios are approximately 2.7, 5.5, 3.8 and 1.6 for Opus~5.5, GPT-5.5,
GPT-6.1 and Sonnet~5.5. Across all 12 model-setting pairs, \rsipde{} uses fewer recorded calls in eight.
The exceptions include three Noisy comparisons: 278 versus 187 calls with Opus~5.5, 539 versus 140 with
GPT-6.1, and 138 versus 117 with Sonnet~5.5. On Hard-OOD, Sonnet~5.5 uses 125 versus 96 calls.
A structured loop therefore changes the allocation of computation, rather than guaranteeing fewer calls.

Dollar cost depends on token lengths, caching and stopping behavior as well as call count.
Full GPT-5.5 costs \$32.01 with \rsipde{} versus \$40.97 with Codex, a 21.9\% reduction using
recorded costs. That measured comparison does not imply lower cost for every model or setting.
The website substitutes a \emph{retrospective stop-rule replay cost} for selected RSI rows: it prices the
calls before the replayed saturation stop, while keeping the full-budget pass rates, calls and trajectories.
These costs, marked $\ast$, are useful estimates of spending that a rule might save, but are not jointly
measured accuracy/cost outcomes from a fresh stopped run. For example, Full Opus~5.5 is listed at \$30.47
under that replay, whereas its 196 passes come from the complete selected run. GPT-6.1 has no dollar
estimate in the current public snapshot. We preserve unavailable values and avoid deriving a universal
cost-efficiency claim from mixed provenance.

\subsection{Residual diagnostics and their interpretation}
\label{sec:residual-results}

Residual-guided identification in \cref{sec:residual-first} uses a hypothesized equation.
The residual table on the website instead evaluates the frozen returned solution under the evaluator's
\emph{true} equation. If its terms are $T_j[\hat u]$, the reported relative residual is
\begin{equation}
 r_{\mathrm{true}}(\hat u)
  = \frac{\|\sum_jT_j[\hat u]\|}{\sum_j\|T_j[\hat u]\|}.
 \label{eq:true-residual}
\end{equation}
This diagnostic needs no reference solution, but does need the true class and coefficients; it is computed
afterward and is never exposed as a hidden-equation selection oracle to the agent.

The calculation uses spectral derivatives in periodic directions, fourth-order differences on bounded grids,
cubic splines for time derivatives and an L1 approximation for Caputo derivatives. Darcy uses a weak residual
against sine modes. The point-cloud elliptic family has no reported residual. These discretizations create a
resolution-dependent floor, so a residual measures equation consistency at the returned resolution, not a
uniform solution-error bound.

\begin{figure}[t]
 \centering
 \includegraphics[width=0.93\textwidth]{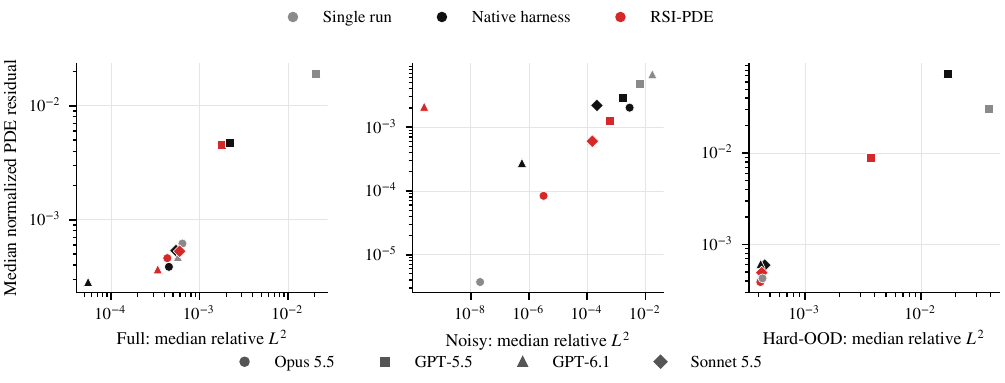}
 \caption{Each point pairs a website row's median relative $L^2$ error with its median post-hoc true-equation
 residual. Points are aggregate medians, not paired per-case observations; marker shapes identify models and
 colors identify harnesses. Coverage and reference runs vary, including historical Full native diagnostics.
 The plot illustrates complementary diagnostics and should not be read as a causal residual-to-accuracy law.}
 \label{fig:residuals}
\end{figure}

For example, the historical Full Opus~5.5 native and RSI rows have median relative $L^2$ errors of roughly
$4.5\times10^{-4}$ and $4.3\times10^{-4}$, but median residuals of $3.9\times10^{-4}$ and
$4.6\times10^{-4}$. A lower aggregate residual therefore need not correspond to a lower aggregate reference
error. The reported rank correlations and passed/failed medians provide additional information, but their
coverage includes only the cases for which the corresponding diagnostics are available.

The Full native GPT-6.1 residual row retains 199/210 reference-run passes, and Sonnet~5.5 retains 193/210,
rather than the updated headline totals 189 and 183. Single-run residual rows can also have different sample
coverage from the headline averages. We label these as reference-run diagnostics in \cref{app:residuals},
preserving the source values instead of presenting them as new measurements of the updated totals.

\section{Discussion and Limitations}
\label{sec:limits}

\paragraph{Unknown physics as a controlled scientific task.}
Jointly hiding equation family and coefficients makes solver construction harder than translating a fully
specified PDE into code. Full, Noisy and Hard-OOD expose different failure modes in that process.
Nevertheless, the task supplies a finite candidate library and known physical inputs. The 1,001-case
collection is mostly synthetic, covers one- and two-dimensional domains, and includes selected data-driven
input fields. It is a reproducible analogue of scientific modeling, not a complete representation of noisy
laboratory data, open-ended equation discovery or industrial workloads. Noisy selection and Hard-OOD
feasibility filtering further restrict the evaluated populations.

\paragraph{Residual guidance and identifiability.}
A wrong equation can have a small residual when observations are sparse, weakly informative or differentiated
inaccurately. Multiple families or parameter values may fit the same samples, especially at 50\% noise.
Keeping five candidates reduces premature commitment but does not guarantee true-family retention. The
reported experiments contain no isolated residual-screening ablation, shortlist-recall measurement or
parameter-recovery rate. Consequently, improved solution pass rates support the complete workflow, without
establishing that residual guidance alone is faster or recovers every true coefficient. Reference-based
$L^2$ checks also inherit interpolation and numerical-reference error; self-convergence does not prove a
uniform reference error bound.

\paragraph{Replication and benchmark reuse.}
Most model--setting comparisons have one run; the three-run Full Opus native range is large relative to
several margins. Repeated seeds and paired initial states are needed for uncertainty estimates and component
ablations. Running agents cannot see hidden test feedback, but maintainers have inspected benchmark results
when selecting design choices, including rules for using known physical functions. The current record is
therefore developmental rather than an untouched holdout estimate \citep{dwork2015reusable,blum2015ladder}.
Noise-level visibility also differs among the reported Noisy runs: GPT-6.1 and Sonnet~5.5 hide it, while
other displayed Noisy rows show it; Hard-OOD hides it throughout. Controlled comparisons should match those
conditions and use fresh committed test draws. Historical Full diagnostics are labeled separately from
updated headline totals.

\paragraph{What evolves and what the resource results mean.}
The model weights and executable supervisor remain fixed, while solver code, policy text and memory can
change. Final-program improvement does not establish that an evolved controller is a better improver;
that claim requires matched continuation experiments. Recorded calls decrease in eight of 12 comparisons,
with substantial exceptions on Noisy. Dollar costs depend on token usage and caching, and selected costs
use retrospective stop-rule replay while retaining full-budget scores. Those estimates require fresh
stopped runs before they can support jointly achieved accuracy/cost claims.

\section{Conclusion}
\label{sec:conclusion}

\bench{} brings together 1,001 PDE/FPDE cases and an unknown-family, unknown-parameter identify-then-solve
task. Full, Noisy and Hard-OOD separately test noise-free accuracy, robustness to severe Gaussian sample
noise, and transfer to unseen templates and shifted coefficients. The \rsipde{} harness uses residuals to
prioritize candidate physical laws, then combines public relative $L^2$ feedback, coefficient refinement
and validation to retain accurate executable solvers and improve their policy and memory. The reported
four-model comparison improves mean pass rates over single-run and native-harness baselines in all three
settings and reduces aggregate recorded model calls. Repeated runs, residual-specific ablations and fresh
hidden evaluations will clarify which parts of this workflow generalize most reliably.

\section*{Data and Code Availability}

The RSI-PDEbench dataset and RSI-PDE implementation are publicly
available at \url{https://github.com/jeremylu916/RSI-PDE}.
The release includes 1,001 PDE and FPDE cases spanning 42
templates, together with case-generation routines, reference
solutions, evaluation code, and the RSI-PDE harness.
The evaluation settings and split-construction procedures are
described in Section~3. The frozen result snapshot and
figure-generation scripts support reproduction of the reported
tables and plots, as detailed in Appendix~D.
Benchmark results and additional documentation are available at
\url{https://jeremylu916.github.io/RSI-PDE/}.

\section*{Acknowledgment}
We gratefully acknowledge the support from DARPA MATHBAC program under grant number: HR0011-26-3-E152, the National Science Foundation (DMS-2533878, DMS-2053746, DMS-2134209, ECCS-2328241, CBET-2347401 and OAC-2311848), and U.S.~Department of Energy (DOE) Office of Science Advanced Scientific Computing Research program under the "Uncertainty Quantification for Multifidelity Operator Learning (MOLUcQ)" project (Project No. 81739), DE-SC0023161, the SciDAC LEADS Institute, and DOE–Fusion Energy Science, under grant number: DE-SC0024583.

\begingroup

\bibliography{rsi_pdebench}
\bibliographystyle{icml2026}
\endgroup

\newpage
\appendix

\section{Related Work}
\label{sec:related}

\paragraph{PDE learning and equation discovery.}
PDEBench, PDEArena, PINNacle and The Well provide important test beds for learned numerical models
\citep{takamoto2022pdebench,gupta2022pdearena,hao2024pinnacle,ohana2024well}. Physics-informed neural networks
use equation residuals in forward and inverse problems \citep{raissi2019pinn}, while neural operators learn
solution mappings across problem instances \citep{li2021fno}. Equation discovery instead seeks a governing law:
sparse identification and PDE-FIND select differential terms from data
\citep{brunton2016sindy,rudy2017pdefind}, and PDE-Net learns differential operators \citep{long2018pdenet}.
\bench{} combines identification with construction of an executable numerical solver. The entrant can use
classical, learned or hybrid methods, but is graded on the returned solution. Five data-driven templates reuse
input fields from DiffusionPDE \citep{huang2024diffusionpde}, with reference solutions recomputed for this benchmark.

\paragraph{Physics-informed numerical learning.}
Training methods address complementary challenges in constructing accurate scientific surrogates.
NSGA-PINN uses multi-objective optimization for physics-informed network training \citep{lu2023nsga},
iPINNER combines iterative physics-informed learning with ensemble Kalman filtering \citep{lu2025ipinner},
and Morephy-Net studies evolutionary multi-objective, replica-exchange physics-informed operator learning
\citep{lu2026morephy}. FPINN-DeepONet targets multi-term time-fractional mixed diffusion--wave equations
\citep{lu2025fpinn}, while Neural-POD develops a neural-operator formulation of nonlinear proper orthogonal
decomposition \citep{mou2026neural}. Spectrally guided Muon optimization and AdamFLIP study optimization
for scientific learning and hard-constrained PINNs \citep{lu2026muonspectral,lu2026adamflip}. These works
provide numerical-learning approaches relevant to the solution methods an agent may investigate. \bench{}
evaluates the broader process of identifying the equation, estimating its coefficients and constructing
a solver under a fixed development budget.

\paragraph{Scientific coding and agent interfaces.}
Scientific and software-engineering benchmarks evaluate programs, research tasks or repository-level changes
\citep{tian2024scicode,jimenez2024swebench,chan2025mlebench,wijk2025rebench}. SWE-agent demonstrates the
importance of the interface through which a model edits and executes code \citep{yang2024sweagent}.
CodePDE studies inference-time generation and refinement of solvers for specified equations
\citep{li2025codepde}. PDEAgent-Bench emphasizes accurate, efficient solver generation for known PDE
specifications and compatibility with multiple FEM libraries \citep{hang2026pdeagentbench}. \bench{}
complements that emphasis by hiding the instance's family and coefficients and including fractional dynamics.
LLM-SR searches for scientific equations represented as programs \citep{shojaee2025llmsr}. Our comparison combines finite-library equation identification with solver development,
keeps the foundation model fixed within a model-specific comparison, and varies the harness under numerical
feedback and a recorded development budget. Unlike a unit-test-only task, a solution must also
meet a quantitative error tolerance over an evaluation grid.

\paragraph{Iterative and self-improving agents.}
Reflexion and Self-Refine use feedback to revise behavior or outputs without retraining model weights
\citep{shinn2023reflexion,madaan2023selfrefine}. Program-search systems use evaluations to improve candidate code
\citep{romeraparedes2024funsearch,novikov2025alphaevolve}, and other systems optimize improvement procedures or
agent designs \citep{zelikman2024stop,hu2025adas}. CONTRAMEM studies self-evolving procedural memory
from contrasting multi-model trajectories \citep{deng2026contramem}, and work on the limits of harness
evolution examines when weight training becomes necessary \citep{tian2026harness}. Muon-OGD studies
spectral orthogonal gradient projection for LLM continual learning \citep{lu2026muon}. In our setting,
the foundation weights remain fixed, and \rsipde{} updates solver code, an instruction policy and memory.
The displayed benchmark runs keep the executable supervisor fixed. Establishing that an evolved controller is
a better improver would require paired continuations from the same solver and memory, beyond a comparison of
final programs. Adaptive reuse of validation or test results also requires care
\citep{dwork2015reusable,blum2015ladder}; we therefore state the provenance of the development results explicitly.

\paragraph{Structured scientific applications and agent reliability.}
Broader scientific-learning applications include SE(3)-MeanFlow for protein-backbone generation on Lie
groups \citep{bai2026se} and MedRouter for routing-based reasoning across medical language models
\citep{cao2026medrouter}. They illustrate structured generation and task-specific routing in domains
beyond PDE solver construction. Agent reliability also has dimensions beyond numerical accuracy:
Chain of Risk studies safety failures and adaptive steering in reasoning models \citep{li2026chainofrisk},
while Think-with-Image studies multimodal jailbreak robustness \citep{tian2026think}. Those questions
complement the execution isolation, information boundaries and quantitative solution checks used here.

\section{Templates}
\label{app:templates}

\cref{tab:templates} lists the 42 templates. In Full, every template has one training case, one validation case and five
hidden-test cases; the other settings have the allocations in \cref{tab:splits}. Templates marked $^\dagger$ were hardened after calibration (\cref{app:calibration}); templates
marked $^\ddagger$ take their input fields from the DiffusionPDE test set.

\begin{table}[!ht]
  \caption{The 42 templates. Dim.: spatial dimension (0 for the fractional ODE). Steady: no time dependence.}
  \label{tab:templates}
  \centering
  \begin{small}
    \begin{tabular}{lllcc}
      \toprule
      ID & Template & Class & Dim. & Steady \\
      \midrule
      \multicolumn{5}{l}{\emph{Classical and fractional templates}} \\
      1a & Poisson, manufactured & Poisson & 1 & yes \\
      1b & Poisson, multiscale & Poisson & 1 & yes \\
      1c & Poisson, interior layer & Poisson & 1 & yes \\
      2a & Heat, manufactured & heat & 1 &  \\
      2b & Heat, free decay & heat & 1 &  \\
      2c & Heat, periodic source & heat & 1 &  \\
      3a & Conv.--diff., manufactured & convection--diffusion & 1 &  \\
      3b$^\dagger$ & Conv.--diff., boundary layer & convection--diffusion & 1 & yes \\
      3c$^\dagger$ & Conv.--diff., layer + source & convection--diffusion & 1 & yes \\
      3d & Conv.--diff., Gaussian pulse & convection--diffusion & 1 &  \\
      4a & Burgers, manufactured & Burgers & 1 &  \\
      4b & Burgers, sine data & Burgers & 1 &  \\
      4c & Burgers, viscous shock & Burgers & 1 &  \\
      5 & Wave 2-D & wave & 2 &  \\
      6$^\dagger$ & Fractional ODE & Caputo ODE & 0 &  \\
      7 & Time-fractional diffusion & Caputo & 1 &  \\
      8 & Multi-term fractional & multi-term Caputo & 1 &  \\
      9$^\dagger$ & Fractional, zero initial data & Caputo & 1 &  \\
      10 & Variable order $\alpha(t)$ & variable order & 1 &  \\
      11$^\dagger$ & Variable order $\alpha(x,t)$ & variable order & 1 &  \\
      \midrule
      \multicolumn{5}{l}{\emph{Nonlinear and multiphysics templates}} \\
      12a & Kuramoto--Sivashinsky, manufactured & Kuramoto--Sivashinsky & 1 &  \\
      12b & Kuramoto--Sivashinsky, chaotic & Kuramoto--Sivashinsky & 1 &  \\
      13a & Navier--Stokes, Taylor--Green & Navier--Stokes & 2 &  \\
      13b & Navier--Stokes, Kovasznay & Navier--Stokes & 2 & yes \\
      14a & Cahn--Hilliard, manufactured & Cahn--Hilliard & 2 &  \\
      14b & Cahn--Hilliard, spinodal & Cahn--Hilliard & 2 &  \\
      15a & Maxwell, cavity mode & Maxwell TE & 2 &  \\
      15b & Maxwell, current-driven & Maxwell TE & 2 &  \\
      16a & Time-harmonic EM, low frequency & curl--curl & 2 & yes \\
      16b & Time-harmonic EM, high frequency & curl--curl & 2 & yes \\
      \midrule
      \multicolumn{5}{l}{\emph{Data-driven and additional-dimensional templates}} \\
      17 & Burgers, Gaussian data$^\dagger$ & Burgers & 1 &  \\
      18 & Burgers, random data$^\ddagger$ & Burgers & 1 &  \\
      19 & Conv.--diff., periodic pulse & convection--diffusion & 1 &  \\
      20 & Darcy flow$^\ddagger$ & Darcy & 2 & yes \\
      21 & Poisson, random forcing$^\ddagger$ & Poisson 2-D & 2 & yes \\
      22 & Helmholtz$^{\dagger\ddagger}$ & Helmholtz & 2 & yes \\
      23 & Navier--Stokes vorticity$^\ddagger$ & vorticity Navier--Stokes & 2 &  \\
      24 & Kolmogorov flow & vorticity Navier--Stokes & 2 &  \\
      25 & Allen--Cahn & Allen--Cahn & 2 &  \\
      26 & Burgers 2-D & Burgers 2-D & 2 &  \\
      27 & Advection--diffusion--reaction & ADR & 2 &  \\
      28 & Elliptic equation & elliptic & 2 & yes \\
      
      \bottomrule
    \end{tabular}
  \end{small}
\end{table}

\FloatBarrier
\section{Reference Construction and Calibration}
\label{app:references}\label{app:calibration}

Analytic references are evaluated directly at the evaluator's nodes. Numerical references are constructed
with a finer computation than the comparison resolution and checked after both are mapped to the common
evaluation grid. Representative settings in the reference builders are:
\begin{itemize}
 \item chaotic Kuramoto--Sivashinsky: 512 Fourier nodes and 120 ETDRK4 substeps per output interval,
       compared with 256 nodes and 60 substeps;
 \item spinodal Cahn--Hilliard: 256 nodes per spatial direction and 100 substeps,
       compared with 128 nodes and 50 substeps;
 \item Gaussian-data Burgers: refined finite-volume computations at 2048 and 1024 spatial nodes;
 \item Poisson and Helmholtz input-field problems: fine/coarse spatial settings of 1020 and 510;
 \item the Elliptic equation reference: a finite-volume discretization with a preconditioned
       conjugate-gradient relative tolerance of $10^{-11}$.
\end{itemize}
The reference scheme and error allowance are fixed by the benchmark, not changed to favor a submitted solver.
Higher order, finer resolution or tighter linear-solver tolerance are used where appropriate to the family;
these settings are not a claim that every reference has the same discretization error.

The public website also includes calibration profiles from fixed probe programs. Its displayed probe
profiles cover 640 variants of 32 templates; the additional multiphysics templates complete the 42-template
benchmark. Calibration informs hardening and the diagnostic Noisy selection. Since subset selection changes
the population being evaluated, the Full, Noisy and Hard-OOD pass rates should not be interpreted as
estimates of the same underlying case distribution.

\section{Implementation and Reproduction}
\label{app:implementation}

\paragraph{Recorded source snapshot.}
The numerical tables and plots are generated from
\path{arxiv/data/website_results_20261107.json}, a frozen copy of the public website after revision
\texttt{76b33da}. The generator is \path{arxiv/make_website_assets.py}.
It does not run an agent or inspect hidden per-case outcomes. The website corrections are presentation-level
provenance events; the original experiment records are retained. The authors clarified that Full denotes the
noise-free \texttt{full0} sampling variant, which sets observation noise to zero. The frozen website snapshot
retains its published \texttt{full} presentation keys and reported totals. This protocol clarification is recorded
in \path{arxiv/data/protocol_corrections.json}; it does not overwrite original experiment records.

\paragraph{Agent-level residuals and supervisor.}
Candidate-equation residuals are implemented by model-written sandbox routines. For example, a retained
fractional solver uses least-squares terms for PDE collocation, sample fit and boundaries before comparing
candidate scores; another solver checks coefficient consistency against known forcing and initial data.
These routines are distinct from evaluator-only true-equation residuals. The top-five policy describes
the identification workflow; the aggregate results do not isolate its effect.

The supervisor is implemented in \path{src/pde_rsi/code_agents.py}; the standard gate is in
\path{src/pde_rsi/evaluator.py}, and the native wrapper in \path{scripts/run_native_agent.py}.
The public-pass rule compares the same public case identifiers before and after an update, gains at least
one pass, loses none, permits no new execution/constraint/resource failure and bounds error changes on
cases that already pass. Model weights and executable control flow stay fixed during the reported runs.
Raw configuration identifiers remain in the result snapshot for reproduction, without defining separate
methods in this paper.

\paragraph{Reproducible paper build.}
The revised arXiv manuscript is maintained directly in this directory, retaining the original document
style and author block. Asset generation uses the frozen
website JSON, and the LaTeX build uses the included style, bibliography and figures.

\FloatBarrier
\section{Additional Learning Curves and Resource Records}
\label{app:results}

\begin{figure}[!ht]
 \centering
 \includegraphics[width=0.9\textwidth]{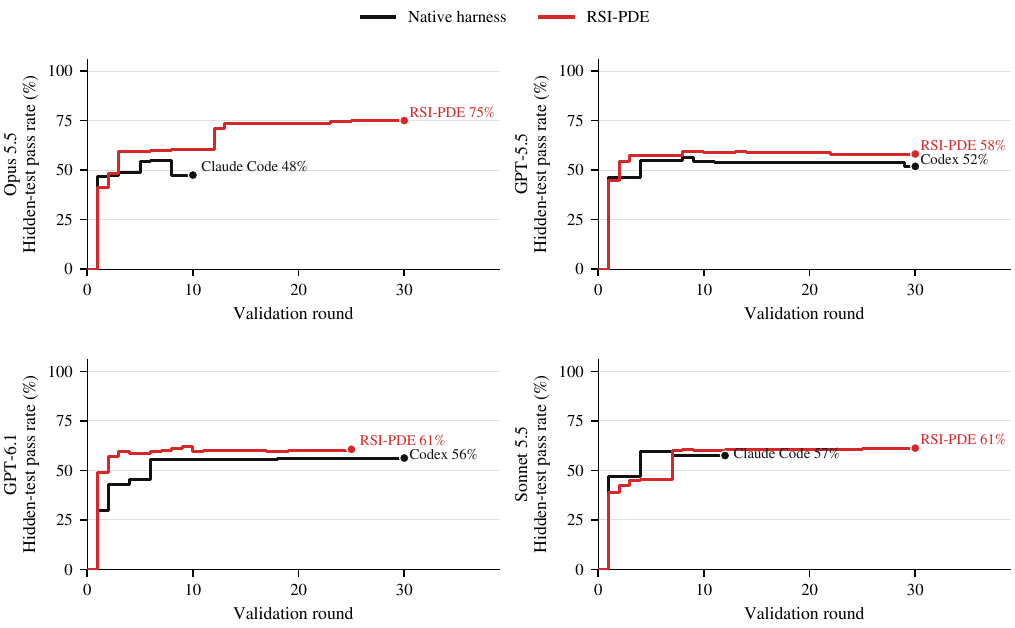}
 \caption{Noisy trajectories for all four models, with corrected GPT-6.1 identities: red \rsipde{} finishes
 at 97/160 and black Codex at 90/160. The curves retain the supplied numerical measurements and change only
 their corrected harness attribution. Round indices are validation-submission indices.}
 \label{fig:curves-noisy}
\end{figure}

\begin{figure}[!ht]
 \centering
 \includegraphics[width=0.9\textwidth]{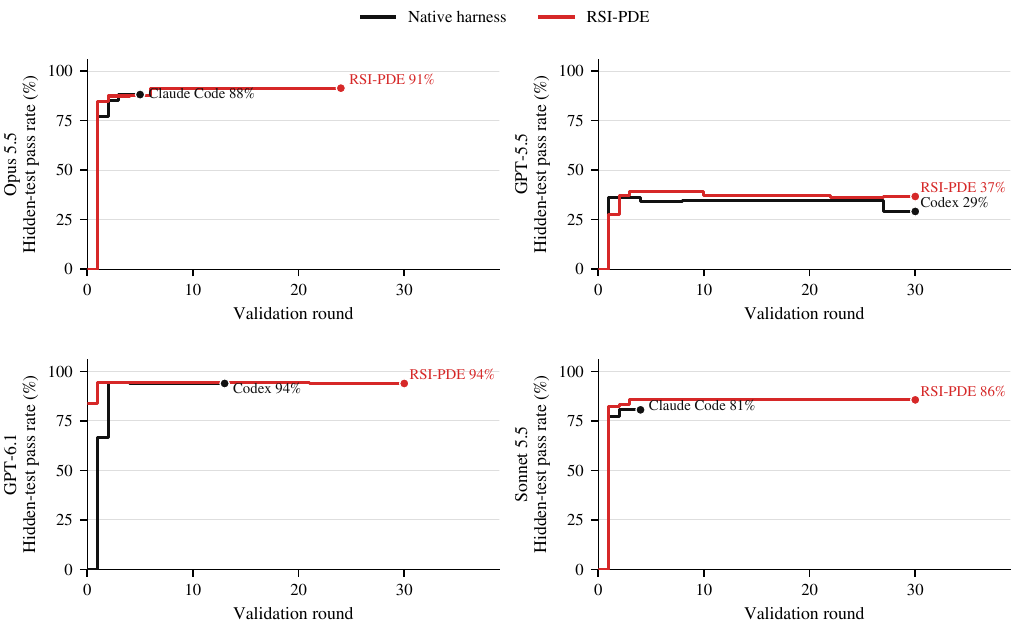}
 \caption{Hard-OOD trajectories. \rsipde{} is red and the native harness is black. Template novelty and
 coefficient shift are both included in the 278-case aggregate; \cref{tab:settings} separates their endpoints.}
 \label{fig:curves-ood}
\end{figure}

\FloatBarrier
\begin{table}[!ht]
 \caption{Website resource records, summed over the development groups and averaged where repeated.
 Recorded rounds are summed group submissions, not the final horizontal coordinate of a combined curve.
 $\dagger$: updated Full native total paired with historical resource records.
 $\ddagger$: corrected Noisy GPT-6.1 attribution.
 $\ast$: retrospective stop-rule replay cost; calls and accuracy remain full-budget values.}
 \label{tab:costs}
 \centering
 \resizebox{0.91\textwidth}{!}{\input{tables/costs.tex}}
\end{table}

\FloatBarrier
\section{Retrospective Residual Tables}
\label{app:residuals}

The following tables reproduce the website's aggregate residual diagnostics. They use the reference-run
pass counts belonging to those diagnostic records. Medians cover cases that returned numerical solution arrays, including solutions that failed the accuracy
test. Residual coverage can be smaller, and point-cloud elliptic outputs have no residual. The last column is
the reported Spearman correlation between available per-case relative $L^2$ errors and residuals, rather
than a correlation computed from the plotted row medians.

\begin{table}[!ht]
 \caption{Full reference-run diagnostics. $\dagger$: historical native counts 199/210 for GPT-6.1 and 193/210
 for Sonnet~5.5; the current headline totals are 189 and 183. Single-run coverage and sample means can also
 differ from the headline comparison. These rows are not new measurements of the updated totals.}
 \label{tab:residual-full}
 \centering
 \resizebox{\textwidth}{!}{\input{tables/residuals_full.tex}}
\end{table}

\begin{table}[!ht]
 \caption{Noisy reference-run diagnostics. The GPT-6.1 harness identities follow the author's correction;
 residual values and their original numerical coverage are preserved.}
 \label{tab:residual-noisy}
 \centering
 \resizebox{\textwidth}{!}{\input{tables/residuals_noisy.tex}}
\end{table}

\begin{table}[!ht]
 \caption{Hard-OOD reference-run diagnostics, including the available single-run rows. The true-equation
 residual is an evaluator-only post-hoc measurement and does not alter the relative-$L^2$ pass criterion.}
 \label{tab:residual-ood}
 \centering
 \resizebox{\textwidth}{!}{\input{tables/residuals_ood.tex}}
\end{table}

\FloatBarrier
\clearpage


\end{document}

%% file: tables/main_full.tex
\begin{tabular}{llrrrrrrr}
\toprule
Model & Entrant & Passed /210 & Rate (\%) & Runs & Gain (pp) & Gap (pp) & Calls & Cost (\$) \\
\midrule
\multirow{3}{*}{Opus 5.5} & Single run & 66 & 31.5 & 4 & -- & -- & 3 & 7.32 \\
 & Claude Code & 193 & 91.9 & 3 & +60.4 & +2.5 & 240 & 16.74 \\
 & RSI-PDE & \textbf{196} & 93.3 & 1 & +61.8 & -0.5 & 88 & 30.47$^{\ast}$ \\
\midrule
\multirow{3}{*}{GPT-5.5} & Single run & 38 & 18.2 & 4 & -- & -- & 3 & 1.22 \\
 & Codex & 95 & 45.2 & 1 & +27.0 & +7.1 & 482 & 40.97 \\
 & RSI-PDE & \textbf{104} & 49.5 & 1 & +31.3 & +5.2 & 88 & 32.01 \\
\midrule
\multirow{3}{*}{GPT-6.1} & Single run & 77 & 37.0 & 4 & -- & -- & 4 & -- \\
 & Codex$^{\dagger}$ & 189 & 90.0 & 1 & +53.0 & +10.0 & 548 & -- \\
 & RSI-PDE & \textbf{193} & 91.9 & 1 & +54.9 & +5.7 & 146 & -- \\
\midrule
\multirow{3}{*}{Sonnet 5.5} & Single run & 59 & 28.1 & 1 & -- & -- & -- & -- \\
 & Claude Code$^{\dagger}$ & 183 & 87.1 & 1 & +59.0 & +3.3 & 148 & 6.41 \\
 & RSI-PDE & \textbf{187} & 89.0 & 1 & +61.0 & +3.8 & 95 & 21.81$^{\ast}$ \\
\bottomrule
\end{tabular}

%% file: tables/settings.tex
\begin{tabular}{llrrrrrr}
\toprule
Model & Entrant & Noisy /160 & Calls & Hard-OOD /278 & Calls & Unseen /188 & Shifted /90 \\
\midrule
\multirow{3}{*}{Opus 5.5} & Single run & 37 & 3 & 133 & 3 & 88& 45 \\
 & Claude Code & 76 & 187 & 245 & 240 & 161 & 84 \\
 & RSI-PDE & \textbf{120} & 278 & \textbf{254} & 81 & \textbf{165} & \textbf{89} \\
\midrule
\multirow{3}{*}{GPT-5.5} & Single run & 51 & 3 & 42 & 3 & 33 & 9\\
 & Codex & 83 & 332 & 81 & 374 & 44 & 37 \\
 & RSI-PDE & \textbf{93} & 104 & \textbf{102} & 91 & \textbf{57} & \textbf{45} \\
\midrule
\multirow{3}{*}{GPT-6.1} & Single run & 45 & 3 & 135 & 3 &  88& 47 \\
 & Codex$^{\ddagger}$ & 90 & 140 & \textbf{261} & 643 & \textbf{175} & 86 \\
 & RSI-PDE$^{\ddagger}$ & \textbf{97} & 539 & \textbf{261} & 111 & 173 & \textbf{88} \\
\midrule
\multirow{3}{*}{Sonnet 5.5} & Single run & 36 & 3 & 104 & 3 & 60 & 44 \\
 & Claude Code & 92 & 117 & 224 & 96 & 134 & \textbf{90} \\
 & RSI-PDE & \textbf{98} & 138 & \textbf{238} & 125 & \textbf{148} & \textbf{90} \\
\bottomrule
\end{tabular}

%% file: tables/costs.tex
\begin{tabular}{lllrrrr}
\toprule
Setting & Model & Entrant & Passed & Calls & Recorded rounds & Cost (\$) \\
\midrule
Full & Opus 5.5 & Single run & 66.3 & 3 & 0 & 7.32 \\
 & Opus 5.5 & Claude Code & 193 & 240 & 20 & 16.74 \\
 & Opus 5.5 & RSI-PDE & 196 & 88 & 46 & 30.47$^{\ast}$ \\
 & GPT-5.5 & Single run & 38.3 & 3 & 0 & 1.22 \\
 & GPT-5.5 & Codex & 95 & 482 & 87 & 40.97 \\
 & GPT-5.5 & RSI-PDE & 104 & 88 & 64 & 32.01 \\
 & GPT-6.1 & Single run & 77.8 & 4 & 0 & -- \\
 & GPT-6.1 & Codex$^{\dagger}$ & 189 & 548 & 20 & -- \\
 & GPT-6.1 & RSI-PDE & 193 & 146 & 67 & -- \\
 & Sonnet 5.5 & Single run & 59 & 3 & 0 & -- \\
 & Sonnet 5.5 & Claude Code$^{\dagger}$ & 183 & 148 & 10 & 6.41 \\
 & Sonnet 5.5 & RSI-PDE & 187 & 95 & 50 & 21.81$^{\ast}$ \\
\midrule
Noisy & Opus 5.5 & Single run & 37 & 3 & 0 & 3.35 \\
 & Opus 5.5 & Claude Code & 76 & 187 & 24 & 13.90 \\
 & Opus 5.5 & RSI-PDE & 120 & 278 & 68 & 67.10$^{\ast}$ \\
 & GPT-5.5 & Single run & 51.5 & 3 & 0 & 1.15 \\
 & GPT-5.5 & Codex & 83 & 332 & 78 & 20.58 \\
 & GPT-5.5 & RSI-PDE & 93 & 104 & 81 & 17.84$^{\ast}$ \\
 & GPT-6.1 & Single run & 45.3 & 3 & 0 & -- \\
 & GPT-6.1 & Codex$^{\ddagger}$ & 90 & 140 & 82 & -- \\
 & GPT-6.1 & RSI-PDE$^{\ddagger}$ & 97 & 539 & 47 & -- \\
 & Sonnet 5.5 & Single run & 36 & 3 & 0 & -- \\
 & Sonnet 5.5 & Claude Code & 92 & 117 & 21 & 4.21 \\
 & Sonnet 5.5 & RSI-PDE & 98 & 138 & 90 & 25.29$^{\ast}$ \\
\midrule
Hard-OOD & Opus 5.5 & Single run & 133 & 3 & 0 & 7.61 \\
 & Opus 5.5 & Claude Code & 245 & 240 & 14 & 18.86 \\
 & Opus 5.5 & RSI-PDE & 254 & 81 & 48 & 17.55$^{\ast}$ \\
 & GPT-5.5 & Single run & 42 & 3 & 0 & 1.06 \\
 & GPT-5.5 & Codex & 81 & 374 & 89 & 22.79 \\
 & GPT-5.5 & RSI-PDE & 102 & 91 & 64 & 22.88 \\
 & GPT-6.1 & Single run & 135 &3 & 0 & -- \\
 & GPT-6.1 & Codex & 261 & 643 & 35 & -- \\
 & GPT-6.1 & RSI-PDE & 261 & 111 & 79 & -- \\
 & Sonnet 5.5 & Single run & 104 & 3 & 0 & -- \\
 & Sonnet 5.5 & Claude Code & 224 & 96 & 10 & 3.33 \\
 & Sonnet 5.5 & RSI-PDE & 238 & 125 & 85 & 10.55$^{\ast}$ \\
\bottomrule
\end{tabular}

%% file: tables/residuals_full.tex
\begin{tabular}{llrrrrrr}
\toprule
Model & Entrant & Ref.\ passes/cases & Median $L^2$ & Residual & Residual, passed & Residual, failed & $\rho$ \\
\midrule
Opus 5.5 & Single run & 68.3/100 & $6.4\!\times\!10^{-4}$ & $6.2\!\times\!10^{-4}$ & $6.2\!\times\!10^{-4}$ & $7.1\!\times\!10^{-4}$ & 0.70 \\
Opus 5.5 & Claude Code & 193/210 & $4.5\!\times\!10^{-4}$ & $3.9\!\times\!10^{-4}$ & $3.4\!\times\!10^{-4}$ & $1.9\!\times\!10^{-3}$ & 0.71 \\
Opus 5.5 & RSI-PDE & 196/210 & $4.3\!\times\!10^{-4}$ & $4.6\!\times\!10^{-4}$ & $4.3\!\times\!10^{-4}$ & $1.5\!\times\!10^{-3}$ & 0.71 \\
GPT-5.5 & Single run & 39.8/190 & $2.1\!\times\!10^{-2}$ & $1.9\!\times\!10^{-2}$ & $7.9\!\times\!10^{-4}$ & $4.3\!\times\!10^{-2}$ & 0.87 \\
GPT-5.5 & Codex & 95/210 & $2.2\!\times\!10^{-3}$ & $4.7\!\times\!10^{-3}$ & $1.2\!\times\!10^{-3}$ & $1.7\!\times\!10^{-2}$ & 0.78 \\
GPT-5.5 & RSI-PDE & 104/210 & $1.8\!\times\!10^{-3}$ & $4.5\!\times\!10^{-3}$ & $7.4\!\times\!10^{-4}$ & $1.5\!\times\!10^{-2}$ & 0.85 \\
GPT-6.1 & Single run & 103.7/116.7 & $5.7\!\times\!10^{-4}$ & $4.7\!\times\!10^{-4}$ & $3.8\!\times\!10^{-4}$ & $1.3\!\times\!10^{-3}$ & 0.77 \\
GPT-6.1 & Codex & 199/210$^{\dagger}$ & $5.5\!\times\!10^{-5}$ & $2.8\!\times\!10^{-4}$ & $2.2\!\times\!10^{-4}$ & $1.9\!\times\!10^{-3}$ & 0.68 \\
GPT-6.1 & RSI-PDE & 193/210 & $3.4\!\times\!10^{-4}$ & $3.7\!\times\!10^{-4}$ & $2.7\!\times\!10^{-4}$ & $2.1\!\times\!10^{-3}$ & 0.77 \\
Sonnet 5.5 & Claude Code & 193/210$^{\dagger}$ & $5.4\!\times\!10^{-4}$ & $5.4\!\times\!10^{-4}$ & $5.2\!\times\!10^{-4}$ & $1.5\!\times\!10^{-3}$ & 0.66 \\
Sonnet 5.5 & RSI-PDE & 187/210 & $6.0\!\times\!10^{-4}$ & $5.3\!\times\!10^{-4}$ & $4.1\!\times\!10^{-4}$ & $2.0\!\times\!10^{-3}$ & 0.73 \\
\bottomrule
\end{tabular}

%% file: tables/residuals_noisy.tex
\begin{tabular}{llrrrrrr}
\toprule
Model & Entrant & Ref.\ passes/cases & Median $L^2$ & Residual & Residual, passed & Residual, failed & $\rho$ \\
\midrule
Opus 5.5 & Single run & 37/42 & $2.1\!\times\!10^{-8}$ & $3.7\!\times\!10^{-6}$ & $2.3\!\times\!10^{-6}$ & $1.3\!\times\!10^{-3}$ & 0.86 \\
Opus 5.5 & Claude Code & 76/160 & $2.8\!\times\!10^{-3}$ & $2.0\!\times\!10^{-3}$ & $6.9\!\times\!10^{-5}$ & $1.3\!\times\!10^{-2}$ & 0.90 \\
Opus 5.5 & RSI-PDE & 120/160 & $3.2\!\times\!10^{-6}$ & $8.4\!\times\!10^{-5}$ & $5.1\!\times\!10^{-5}$ & $1.2\!\times\!10^{-2}$ & 0.89 \\
GPT-5.5 & Single run & 51.5/160 & $6.3\!\times\!10^{-3}$ & $4.8\!\times\!10^{-3}$ & $1.6\!\times\!10^{-3}$ & $1.5\!\times\!10^{-2}$ & 0.76 \\
GPT-5.5 & Codex & 83/160 & $1.6\!\times\!10^{-3}$ & $2.9\!\times\!10^{-3}$ & $5.5\!\times\!10^{-4}$ & $1.5\!\times\!10^{-2}$ & 0.92 \\
GPT-5.5 & RSI-PDE & 93/160 & $6.0\!\times\!10^{-4}$ & $1.3\!\times\!10^{-3}$ & $2.3\!\times\!10^{-4}$ & $1.1\!\times\!10^{-2}$ & 0.84 \\
GPT-6.1 & Single run & 45.3/124.3 & $1.7\!\times\!10^{-2}$ & $6.8\!\times\!10^{-3}$ & $8.7\!\times\!10^{-5}$ & $2.1\!\times\!10^{-2}$ & 0.94 \\
GPT-6.1 & Codex & 90/160 & $5.7\!\times\!10^{-7}$ & $2.8\!\times\!10^{-4}$ & $5.6\!\times\!10^{-5}$ & $3.1\!\times\!10^{-2}$ & 0.93 \\
GPT-6.1 & RSI-PDE & 97/160 & $2.5\!\times\!10^{-10}$ & $2.1\!\times\!10^{-3}$ & $4.3\!\times\!10^{-4}$ & $9.4\!\times\!10^{-3}$ & 0.79 \\
Sonnet 5.5 & Claude Code & 92/160 & $2.1\!\times\!10^{-4}$ & $2.2\!\times\!10^{-3}$ & $6.0\!\times\!10^{-5}$ & $1.1\!\times\!10^{-2}$ & 0.91 \\
Sonnet 5.5 & RSI-PDE & 98/160 & $1.5\!\times\!10^{-4}$ & $6.0\!\times\!10^{-4}$ & $5.4\!\times\!10^{-5}$ & $1.3\!\times\!10^{-2}$ & 0.96 \\
\bottomrule
\end{tabular}

%% file: tables/residuals_ood.tex
\begin{tabular}{llrrrrrr}
\toprule
Model & Entrant & Ref.\ passes/cases & Median $L^2$ & Residual & Residual, passed & Residual, failed & $\rho$ \\
\midrule
Opus 5.5 & Single run & 133/133 & $4.3\!\times\!10^{-4}$ & $4.3\!\times\!10^{-4}$ & $4.3\!\times\!10^{-4}$ & -- & 0.60 \\
Opus 5.5 & Claude Code & 245/278 & $4.2\!\times\!10^{-4}$ & $4.1\!\times\!10^{-4}$ & $3.4\!\times\!10^{-4}$ & $2.4\!\times\!10^{-2}$ & 0.66 \\
Opus 5.5 & RSI-PDE & 254/278 & $4.1\!\times\!10^{-4}$ & $3.9\!\times\!10^{-4}$ & $3.6\!\times\!10^{-4}$ & $3.6\!\times\!10^{-2}$ & 0.62 \\
GPT-5.5 & Single run & 42/278 & $3.8\!\times\!10^{-2}$ & $3.0\!\times\!10^{-2}$ & $7.7\!\times\!10^{-4}$ & $5.3\!\times\!10^{-2}$ & 0.82 \\
GPT-5.5 & Codex & 81/278 & $1.7\!\times\!10^{-2}$ & $7.3\!\times\!10^{-2}$ & $4.6\!\times\!10^{-4}$ & $1.5\!\times\!10^{-1}$ & 0.87 \\
GPT-5.5 & RSI-PDE & 102/278 & $3.7\!\times\!10^{-3}$ & $8.8\!\times\!10^{-3}$ & $6.0\!\times\!10^{-4}$ & $1.0\!\times\!10^{-1}$ & 0.90 \\
GPT-6.1 & Codex & 261/278 & $4.2\!\times\!10^{-4}$ & $6.1\!\times\!10^{-4}$ & $5.7\!\times\!10^{-4}$ & $3.8\!\times\!10^{-3}$ & 0.38 \\
GPT-6.1 & RSI-PDE & 261/278 & $4.2\!\times\!10^{-4}$ & $4.6\!\times\!10^{-4}$ & $4.2\!\times\!10^{-4}$ & $1.4\!\times\!10^{-2}$ & 0.62 \\
Sonnet 5.5 & Claude Code & 224/278 & $4.5\!\times\!10^{-4}$ & $6.0\!\times\!10^{-4}$ & $4.5\!\times\!10^{-4}$ & $2.2\!\times\!10^{-2}$ & 0.65 \\
Sonnet 5.5 & RSI-PDE & 238/278 & $4.2\!\times\!10^{-4}$ & $4.9\!\times\!10^{-4}$ & $4.6\!\times\!10^{-4}$ & $5.8\!\times\!10^{-2}$ & 0.59 \\
\bottomrule
\end{tabular}